\documentclass{article}

\usepackage{microtype}
\usepackage{graphicx}
\usepackage{subcaption}
\usepackage{booktabs} % for professional tables

\usepackage{hyperref}

\usepackage[preprint]{icml2026}

\usepackage{amsmath}
\usepackage{amssymb}
\usepackage{mathtools}
\usepackage{amsthm}

\usepackage[capitalize,noabbrev]{cleveref}

\theoremstyle{plain}

\theoremstyle{definition}

\theoremstyle{remark}

\usepackage[textsize=tiny]{todonotes}

\icmltitlerunning{Submission and Formatting Instructions for ICML 2026}

\begin{document}

\twocolumn[
  \icmltitle{Can AI Evaluate AI Scientists?
    A Benchmarking Study of Autonomous Research Generation Systems Using Automated Multi-Model Review}

  \icmlsetsymbol{equal}{*}

  \begin{icmlauthorlist}
    \icmlauthor{Vaibhava Lakshmi Ravideshik}{aff1,equal}
    \icmlauthor{Mayank Kejriwal}{aff1,equal}
  \end{icmlauthorlist}

  \icmlaffiliation{aff1}{GRAIL}

  \icmlcorrespondingauthor{Vaibhava Lakshmi Ravideshik}{vaibhava@grail.page}

  % You may provide any keywords that you find helpful for describing your
  % paper; these are used to populate the "keywords" metadata in the PDF but
  % will not be shown in the document
  \icmlkeywords{AI scientist systems, automated paper generation, automated evaluation, benchmarking, large language models}

  \vskip 0.3in
]

% this must go after the closing bracket ] following \twocolumn[ ...

% This command actually creates the footnote in the first column listing the
% affiliations and the copyright notice. The command takes one argument, which
% is text to display at the start of the footnote. The \icmlEqualContribution
% command is standard text for equal contribution. Remove it (just {}) if you
% do not need this facility.

% Use ONE of the following lines. DO NOT remove the command.
% If you have no special notice, KEEP empty braces:
\printAffiliationsAndNotice{}  % no special notice (required even if empty)
% Or, if applicable, use the standard equal contribution text:
% \printAffiliationsAndNotice{\icmlEqualContribution}

\begin{abstract}
AI Scientist systems capable of autonomous research have the potential to significantly accelerate scientific discovery. However, evaluating and comparing the quality of AI-generated papers remains an open challenge. We propose and implement a rigorous benchmarking protocol using an automated peer-review system that harnesses frontier large language models to assess scientific papers across four core dimensions: originality, scientific rigor, clarity, and significance. We evaluate four leading AI Scientist frameworks: \textit{Sakana AI (v1 \& v2)}, \textit{CycleResearcher}, and \textit{Data-to-Paper}. Each framework was run on a consistent set of 15 research proposals published by a commercial autonomous AI scientist company (FARS), generating 60 papers that we evaluate alongside 15 FARS benchmark papers. Using three independent LLM reviewers (GPT-5.4, Gemini, and Claude), we find that FARS benchmark papers significantly outperform all competing frameworks, achieving mean scores of 2.14--2.47 on a 1--5 scale compared to 1.00--1.87 for other systems. Notably, FARS scores are more than 2$\times$ higher than the next-best systems on Gemini and Claude evaluations. We find strong agreement among Gemini and Claude ($\rho$ = 0.907, $p < 0.001$), and both correlate extremely strongly with the synthesis score ($\rho$ = 0.961, $p < 0.001$), validating the reliability of automated evaluation. However, GPT-5.4 exhibits weaker agreement ($\rho \approx 0.32$), suggesting it evaluates papers using different criteria. These results establish the first quantitative benchmark for AI Scientist systems and demonstrate that multi-model LLM evaluation provides a scalable, consistent framework for assessing autonomous research quality.
\end{abstract}

\section{Introduction}

Automating scientific research represents a significant frontier in contemporary AI. Large language models have recently enabled the development of AI Scientist systems: automated pipelines capable of generating complete research papers from problem proposals, handling hypothesis generation, experimentation, and manuscript composition with minimal human oversight. Recent work including the AI Scientist \cite{Lu2024AIScientistvX} and its successor AI Scientist v2 \cite{Yamada2025AIScientistv2}, along with systems like CycleResearcher \cite{Weng2024CycleResearcher}, Data-to-Paper \cite{Technion2024DataToPaper}, and Agent Laboratory \cite{Schmidgall2025AgentLab}, demonstrate the potential of end-to-end research automation. These systems promise to dramatically accelerate discovery by democratizing access to research capabilities and scaling the pace of knowledge production. However, rapid proliferation of these systems has outpaced our ability to rigorously evaluate them. Despite their growing adoption and deployment, fundamental questions remain unanswered: How do different AI Scientist frameworks compare in output quality? Which systems produce papers closest to established quality standards? Can we systematically identify the strengths and weaknesses of each approach?

The challenge of benchmarking AI-generated research differs fundamentally from traditional performance evaluation. Unlike machine learning systems evaluated on fixed datasets, AI Scientist systems must be assessed on their end-to-end ability to conduct plausible, rigorous, and novel research. This requires evaluating multiple dimensions simultaneously: the originality of hypotheses, the soundness of experimental design, the validity of conclusions, and the clarity of presentation. Traditional human peer review, while gold-standard, is prohibitively expensive at scale. Automated evaluation systems could enable rapid iteration and improvement \cite{Jin2024AgentReview}, but creating trustworthy automated reviewers for scientific papers remains an open problem. Recent work on agent benchmarking \cite{Huang2024MLAgentBench} highlights the complexity of evaluating open-ended autonomous systems. Furthermore, the computational cost and time required to execute and evaluate even a few AI Scientist systems on the same set of proposals has prevented systematic comparison studies.

To address these challenges, we conduct a rigorous end-to-end comparative benchmarking study evaluating four leading autonomous AI Scientist systems: Sakana AI v1, Sakana AI v2, CycleResearcher \cite{Weng2024CycleResearcher}, and Data-to-Paper \cite{Technion2024DataToPaper}. Each system is executed on the same set of 15 research proposals from the FARS (Fully Automated Research System) dataset, generating 60 papers whose quality we assess using our multi-model evaluation framework. For reference, we also evaluate papers generated by FARS itself on the same proposals, creating a controlled experimental design that enables fair comparison across systems with different architectures and input requirements. The resulting dataset comprises 75 papers (60 from the four systems and 15 FARS benchmarks) evaluated systematically across multiple dimensions, enabling comprehensive analysis of relative strengths and weaknesses.

To enable rigorous comparative evaluation at scale, we employ a multi-model evaluation framework that harnesses three frontier large language models-GPT 5.4, Claude Opus 4.6, and Gemini 3.1 Pro-as independent reviewers. Each model independently assesses papers across four core evaluation dimensions: Originality (novelty of contributions), Scientific Rigor (soundness of methodology and evidence), Clarity (quality of exposition and organization), and Significance (importance and potential impact). Each dimension is scored on a 1--5 scale by all three models, with results synthesized into a unified consensus assessment. This multi-model approach provides both quantitative scores and qualitative analysis, including written summaries of paper strengths and weaknesses, identified limitations, and specific recommendations for improvement. Critically, this evaluation process completes within 15--30 minutes per paper, making systematic comparison of multiple systems feasible for the first time.

{Specific contributions that we make in this paper are as follows:} (i) We conduct the \textbf{first rigorous comparative benchmark study} of major AI Scientist systems, evaluating four leading frameworks on identical research proposals and comparing their outputs against a high-quality reference standard (FARS). (ii) We introduce a practical, scalable framework for systematically assessing AI-generated research papers using multiple frontier LLMs as independent reviewers across four core evaluation dimensions.
  (iii) We provide \textbf{quantitative evidence of performance gaps} between systems, demonstrating that FARS benchmark papers significantly outperform papers generated by competing frameworks (over 2$\times$ higher scores across multiple reviewer models). (iv) We demonstrate that \textbf{AI-based evaluation can effectively and consistently assess research quality}, providing both quantitative scores and qualitative feedback, and opening new possibilities for automated iteration and improvement of autonomous research systems.

% The remainder of this paper is organized as follows. Section~2 provides detailed background on the four AI Scientist frameworks and the FARS benchmark system. Section~3 describes GRAIL Reviewer and our evaluation methodology. Section~4 presents the experimental setup and execution. Section~5 reports quantitative results comparing framework performance and reviewer agreement. Finally, Section~6 discusses implications of these findings for the development and deployment of autonomous research systems.

% This paper addresses this gap by introducing GRAIL Reviewer, a multi-agent evaluation system that leverages frontier LLMs to provide rigorous, scalable assessment of research quality. By conducting a controlled benchmarking study across four major frameworks operating on identical research proposals, we provide the first quantitative evidence of performance variations among production autonomous research systems and establish a reproducible framework for future evaluation efforts.

\section{Evaluated AI Scientists and Benchmark Reference (FARS)}

\begin{table*}[h]
\centering
\caption{Architectural and methodological characteristics of the four evaluated AI Scientist systems and the FARS benchmark reference.}
\small
\setlength{\tabcolsep}{2pt}
\begin{tabular}{l@{\hspace{0.3em}}c@{\hspace{0.3em}}c@{\hspace{0.3em}}c@{\hspace{0.3em}}c@{\hspace{0.3em}}c}
\toprule
\textbf{Characteristic} & \textbf{Sakana v1} & \textbf{Sakana v2} & \textbf{CycleR.} & \textbf{Data-to-Paper} & \textbf{FARS} \\
\midrule
Pipeline Design & Linear 4-stage & Agentic tree search & Iterative cycle & Data-driven & Multi-agent \\
Input Type & Problem spec & Problem spec & Problem spec & Dataset & Problem spec \\
Domain Coverage & Limited & Domain-agnostic & Data-agnostic & Data-centric & Domain-aware \\
Integrated Review & No & No & Yes & No & Yes \\
Key Mechanisms & Sequential exec. & VLM feedback + backtracking & RL training & Statistical rigor & Shared workspace \\
\bottomrule
\end{tabular}
\label{tab:framework-comparison}
\end{table*}

This benchmarking study compares four leading autonomous AI Scientist systems against a reference standard by subjecting their outputs to rigorous evaluation on identical research proposals. Table~\ref{tab:framework-comparison} provides an overview of the architectural and methodological characteristics of each system. This section describes the four systems under evaluation: Sakana AI (v1 and v2), CycleResearcher, and Data-to-Paper, along with FARS (Fully Automated Research System), which serves as our benchmark reference. Understanding the design philosophy and distinctive capabilities of each framework is essential for interpreting the comparative performance results presented subsequently.

\subsection{FARS: Benchmark Reference System}
\label{sec:fars}

FARS (Fully Automated Research System) is a multi-agent system designed to autonomously perform the complete research workflow without human intervention. The four specialized agents (Ideation Agent for literature review and hypothesis generation; Planning Agent for experimental methodology; Experiment Agent for experiment execution with access to 160 NVIDIA GPUs; Writing Agent for paper generation) coordinate through a shared file system serving as workspace and persistent memory. FARS focuses on single, well-scoped contributions that may include negative results, representing an important instantiation of autonomous research at scale. Despite limitations in compute-intensive capabilities and human involvement requirements, FARS papers serve as a quality reference standard for evaluating other AI Scientist frameworks.

\subsection{Sakana AI -- The AI Scientist (v1)}
\label{sec:sakana-v1}

The AI Scientist v1 system \cite{Lu2024AIScientistvX} constitutes an end-to-end automated research pipeline implementing a fully autonomous agent architecture. The system accepts a research problem specification as input and generates a complete research manuscript as output. The pipeline comprises four principal stages: \textit{Idea Generation}, which leverages large language models to synthesize novel research hypotheses and methodological directions from the problem description and initial codebase; \textit{Experimentation}, wherein the system automatically generates and executes Python implementations of proposed experiments, including model training, baseline comparisons, and iterative refinement based on empirical results; \textit{Analysis and Visualization}, which performs quantitative analysis of findings and automated generation of figures and tables; and \textit{Manuscript Composition}, which produces a complete research document in LaTeX format, encompassing abstract, introduction, methodology, experimental results, and bibliographic references.

% This framework demonstrates the feasibility of executing the complete research workflow without human intervention, establishing a foundational architecture for fully automated scientific discovery pipelines.

\subsection{Sakana AI -- The AI Scientist (v2)}
\label{sec:sakana-v2}

The AI Scientist v2 \cite{Yamada2025AIScientistv2} advances upon the v1 architecture by introducing several methodological enhancements that address prior limitations. In place of a linear sequential pipeline, v2 employs \textit{Agentic Tree Search} to systematically explore the space of research hypotheses and experimental directions. This approach enables the system to evaluate multiple promising branches in parallel and execute strategic backtracking when experiments yield limited evidence or negative results. Furthermore, v2 achieves \textit{Domain Generalization}, eliminating the requirement for human-curated code templates and enabling the system to operate across diverse research domains without domain-specific customization. Additionally, the framework incorporates \textit{Vision-Language Model-Based Feedback Mechanisms}, whereby generated visualizations are evaluated by multimodal language models to optimize clarity and visual presentation, with iterative refinement of figures and plots guided by model feedback.

The v2 framework shows substantial improvements in output quality. Papers generated by the system were accepted at peer-reviewed workshops at the International Conference on Learning Representations, with peer review scores matching the median distribution of human-authored submissions. This indicates that the system produces research outputs competitive with professional researchers.

\subsection{CycleResearcher}
\label{sec:cycleresearcher}

CycleResearcher \cite{Weng2024CycleResearcher} instantiates an iterative framework that tightly integrates research synthesis with automated quality assessment through two complementary agents operating within a feedback loop: the \textit{Researcher Agent}, which executes the full spectrum of research activities (systematic literature review, manuscript synthesis, and iterative refinement), and the \textit{Reviewer Agent}, which models peer review by analyzing manuscripts and generating detailed critiques. The framework leverages reinforcement learning trained on curated datasets of peer reviews (Review-5k) and scientific literature (Research-14k) to align both agents with authentic scientific standards. CycleResearcher achieves an average quality score of 5.36 on a standardized 5-point scale, which matches human preprints (5.24) and approaches accepted conference papers (5.69).

\subsection{Data-to-Paper}
\label{sec:data-to-paper}

Data-to-Paper \cite{Technion2024DataToPaper} adopts a fundamentally distinct approach that prioritizes empirical data as the primary input to the research discovery process. Rather than following the conventional paradigm of problem formulation followed by experimental design, this system accepts a dataset as input and implements a data-centric workflow. The pipeline comprises three principal stages: \textit{Exploratory Data Analysis}, wherein the system conducts comprehensive statistical and visual examination to identify patterns, correlations, and distributional properties within the data; \textit{Hypothesis Generation}, wherein salient empirical patterns inform the formulation of testable hypotheses; and \textit{Statistical Validation}, wherein candidate hypotheses are subjected to rigorous statistical testing with appropriate methodological controls. Finally, the system produces \textit{Manuscript Generation}, synthesizing validated findings into a coherent research narrative.

The Data-to-paper system is particularly well-suited to discovery-oriented research contexts in which rich datasets provide the foundation for scientific inquiry, emphasizing methodological rigor in statistical inference and data fidelity throughout the synthesis process.

% \subsection{Framework Comparison}

\section{Experimental Setup and Protocol}

\begin{figure*}
  \centering
  \includegraphics[width=0.75\textwidth]{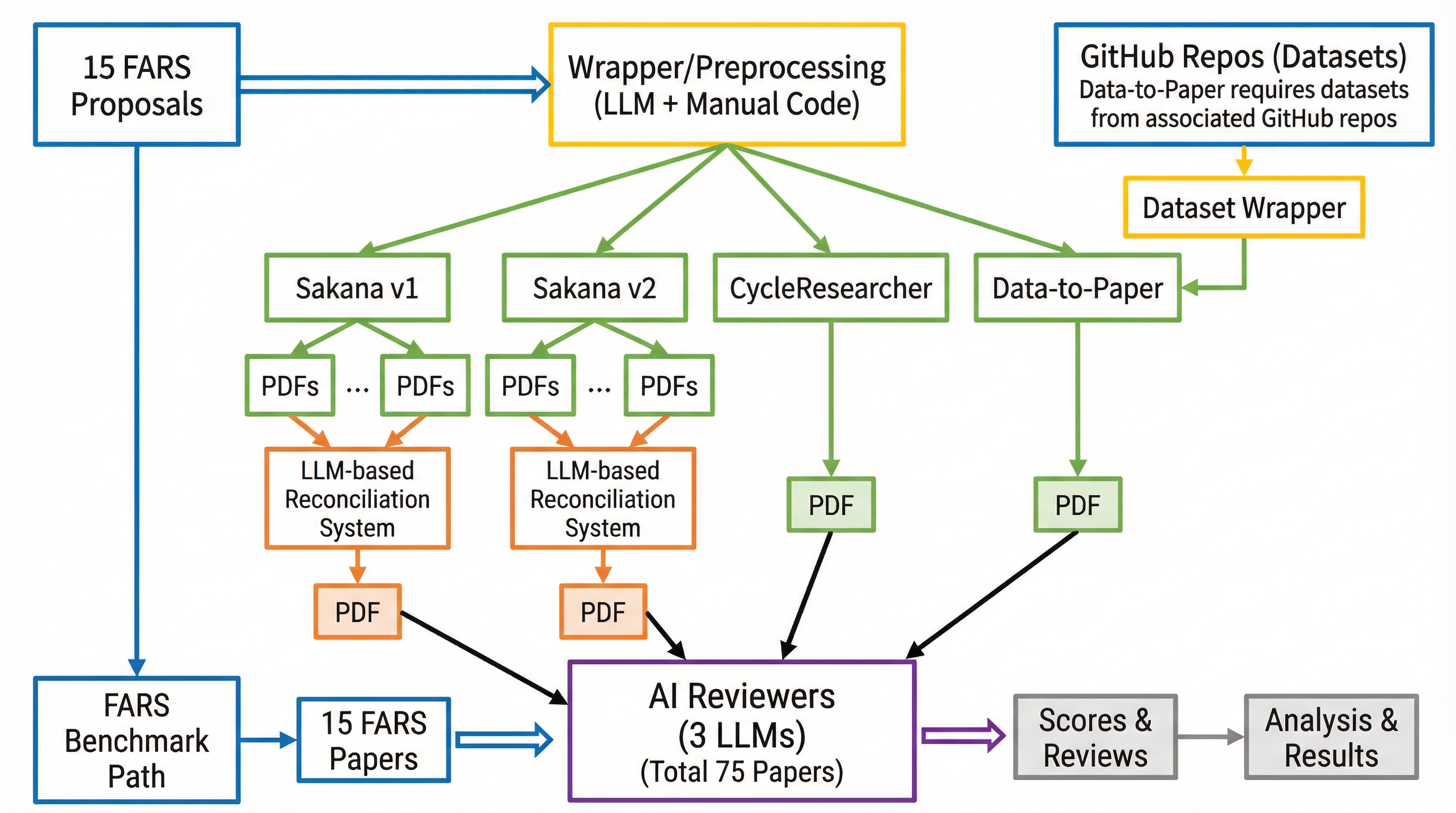}
  \caption{Experimental protocol workflow, illustrating the complete pipeline for evaluating the four AI Scientist frameworks. Starting from 15 FARS proposals, custom wrappers convert proposals into framework-specific inputs (with Data-to-Paper receiving datasets from associated GitHub repositories instead of proposals). Each of the four systems processes its respective input in parallel, generating PDF outputs. Sakana v1 and v2 produce multiple PDFs per proposal, which are merged using an LLM-based reconciliation system. All resulting papers (60 from frameworks + 15 FARS benchmarks = 75 total) are evaluated by the AI Reviewers (three independent LLMs), producing scores, reviews, and final analysis.}
  \label{fig:protocol}
\end{figure*}

% \subsection{Dataset: FARS Proposals and Benchmark Papers}
Fo our benchmarking study, we use a curated set of 15 research proposals generated by the FARS (Fully Automated Research System) system\footnote{\url{https://analemma.ai/fars/}}. These proposals specify research directions in AI safety and machine learning, including detailed background, motivation, and expected outcomes. Notably, the selection of these 15 proposals was guided by the requirements of Data-to-Paper, which requires datasets as input rather than problem specifications. Consequently, we selected proposals for which FARS had released associated GitHub repositories containing empirical datasets. This constraint ensured that all four frameworks could be evaluated on a consistent set of proposals while accommodating each system's distinct input requirements.

For each of these 15 proposals, FARS itself generated a complete research paper using its multi-agent pipeline. These 15 FARS-generated papers serve as benchmark references for comparison, representing a quality standard produced by a mature automated research system. Our evaluation compares papers generated by the four frameworks (Sakana v1, Sakana v2, CycleResearcher, Data-to-Paper) against these FARS benchmark papers on the same proposals, providing a controlled comparison.

\subsection{Paper Generation}
All four frameworks (Sakana v1, Sakana v2, CycleResearcher, and Data-to-Paper) were run on each of the 15 proposals independently. To ensure standardized input across heterogeneous systems, we developed custom input wrappers for each framework that accommodate their distinct input specifications. For Sakana v1, Sakana v2, and CycleResearcher, wrappers convert FARS proposals into their respective system-specific formats: converting proposal specifications into appropriate problem formulations, structuring background information and requirements, and preparing code repositories as needed. For Data-to-Paper, which requires datasets rather than proposals, the wrapper extracts and preprocesses the empirical datasets from the associated GitHub repositories, preparing them in the format expected by the system. Each framework was then provided with its converted input and produced research output.

For CycleResearcher and Data-to-Paper, each framework produced one complete research paper per proposal. However, Sakana v1 and Sakana v2 generate multiple papers corresponding to different research ideas or hypotheses. We configured both Sakana v1 and Sakana v2 to generate 3 ideas per proposal. To produce a single paper output per proposal from these systems, we merged the individual LaTeX files generated by Sakana v1 and v2 using an LLM. The LLM received the raw LaTeX outputs from all ideas and produced a consolidated paper that integrates the most significant contributions and findings across the multiple ideas into a unified narrative.

This processing resulted in 75 papers for review: 15 papers from Sakana v1 (merged from 45 idea-specific papers: 15 proposals × 3 ideas), 15 papers from Sakana v2 (merged from 45 idea-specific papers: 15 proposals × 3 ideas), 15 papers from CycleResearcher, 15 papers from Data-to-Paper, and 15 AI-generated FARS benchmark papers. In total, 60 papers came from the four frameworks and 15 from the FARS benchmark.

\subsection{Evaluation Protocol}

We employ a multi-model evaluation framework to assess research paper quality with rigor and consistency. Our approach leverages a comprehensive automated evaluation system that examines papers across four core dimensions: \textit{Structural Coherence} (presence and quality of standard paper sections with logical narrative flow), \textit{Citation Analysis} (appropriateness of citations and engagement with relevant prior work), \textit{Novelty Assessment} (identification of genuinely novel contributions versus incremental claims), and \textit{Methodological Rigor} (soundness of experimental design, statistical validity, and appropriateness of baselines). We supplement this with an independent LLM-based baseline evaluator that assesses papers across four complementary dimensions: \textit{Originality}, \textit{Scientific Rigor}, \textit{Clarity}, and \textit{Significance}. Both evaluation systems rate papers on a 1--5 scale, allowing us to examine both the robustness of automated evaluation and the consistency of quality signals across different assessment approaches.

All four frameworks (Sakana v1, Sakana v2, CycleResearcher, and Data-to-Paper) were executed independently on each of the 15 FARS proposals. Execution parameters were held constant across frameworks to ensure fair comparison, and each framework produced a complete paper in PDF or LaTeX format without human intervention. Generated papers ranged from 6 to 15 pages in length. This controlled setup produced a total of 60 papers from the four systems (15 per system), which we evaluated alongside 15 FARS benchmark papers on identical proposals.

For each paper, both evaluation systems were applied to generate comprehensive quality assessments. For each framework and evaluation dimension, we computed average scores across its 15 papers to assess central tendency, standard deviation to quantify consistency of output quality, and comparisons to FARS benchmark paper scores to establish relative performance levels. We conducted pairwise framework comparisons using statistical tests to identify significant differences, and analyzed agreement between the two independent evaluation systems by computing Spearman rank correlation on their overall scores. This multi-faceted analysis enables us to understand not only which systems perform best, but also whether automated evaluation provides reliable and consistent quality signals suitable for iterative improvement of AI-generated research.

\section{Results and Discussion}

\begin{figure*}[h!]
\centering
\includegraphics[width=0.95\textwidth]{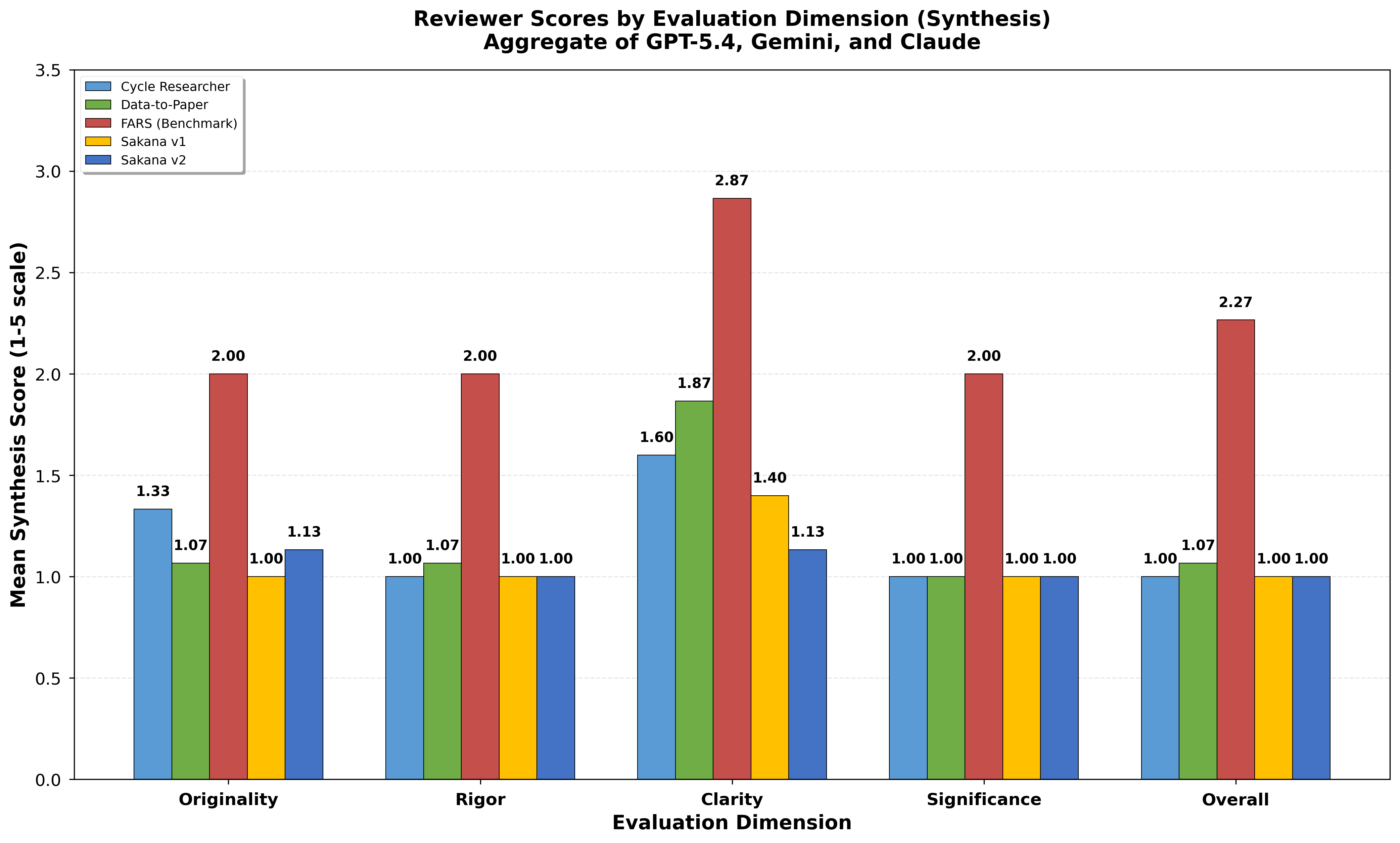}
\caption{Synthesis scores (aggregate across GPT-5.4, Gemini, and Claude reviewers) broken down by evaluation dimension. FARS excels across all four dimensions-Originality (2.00), Rigor (2.00), Clarity (2.87), and Significance (2.00)-while competing systems cluster near the minimum. Cycle Researcher shows the best performance among non-FARS systems, particularly in Clarity (1.60). Error bars indicate standard deviation.}
\label{fig:dimensions}
\end{figure*}

\textbf{Finding 1: FARS significantly outperforms all competing frameworks, with consistent gains across all evaluation dimensions and reviewer models.} The proposal-level evaluation scores presented in Table~\ref{tab:proposal-level-scores} demonstrate that FARS benchmark papers achieve substantially higher ratings compared to the four AI Scientist frameworks across all four reviewer models (GPT-5.4: 2.14, Gemini: 2.47, Claude: 2.47, Synthesis: 2.27). Notably, FARS scores exceed all competing systems by more than 2$\times$ on Gemini and Claude evaluations, indicating a substantial quality gap. This performance advantage is consistently visualized across all evaluation dimensions in Figure~\ref{fig:dimensions}, where FARS demonstrates superior performance in Originality, Rigor, Clarity, and Significance. Among competing frameworks, CycleResearcher exhibits the strongest relative performance, particularly in Clarity (1.60), though it remains substantially below the FARS benchmark across all dimensions.

\begin{table*}[t]
\centering
\caption{Reviewer scores for all 15 proposals across 5 AI Scientist systems. Each cell shows GPT-5.4 / Gemini / Claude / Synthesis overall scores (1--5 scale).}
\tiny
\setlength{\tabcolsep}{1pt}
\begin{tabular}{l|ccccc}
\toprule
\textbf{ID} \hfill \textbf{Title} \hfill & \textbf{CycleRes} & \textbf{Data2Paper} & \textbf{FARS} & \textbf{SakanaV1} & \textbf{SakanaV2} \\
\midrule
FA0001 Canary Control for Emergent Misalignment & 2/1/1/1 & 2/1/1/1 & 2/2/3/2 & 2/1/1/1 & 2/1/1/1 \\
FA0005 Mechanistic Interpretability via Sparse Autoencoders & 2/1/1/1 & 2/1/1/1 & 2/3/3/3 & 2/1/1/1 & 2/1/1/1 \\
FA0006 Web-Agent Evaluation with Evidence-Based Judging & 2/1/1/1 & 1/1/1/1 & 3/4/3/3 & 2/1/1/1 & 2/1/1/1 \\
FA0007 Subsegment Scanning for Padding Attacks & 2/1/1/1 & 2/1/1/1 & 3/3/3/3 & 2/1/1/1 & 1/1/1/1 \\
FA0013 Recurrent VLA Policy Stability Analysis & 2/1/1/1 & 2/2/1/2 & 2/2/2/2 & 1/1/1/1 & 1/1/1/1 \\
FA0015 Reward Model Overoptimization in Jailbreaks & 2/1/1/1 & 1/1/1/1 & 2/2/2/2 & 2/1/1/1 & 2/1/1/1 \\
FA0020 Code Execution Verification for Hallucination Reduction & 2/1/1/1 & 2/1/1/1 & 2/2/2/2 & 2/1/1/1 & 2/1/1/1 \\
FA0021 Concept Drift in Domain-Transfer Fine-tuning & 2/1/1/1 & 2/1/1/1 & 2/2/2/2 & 2/1/1/1 & 2/1/1/1 \\
FA0035 Uncertainty Calibration via Ensemble Disagreement & 2/1/1/1 & 2/1/1/1 & 2/4/2/2 & 2/1/1/1 & 1/1/1/1 \\
FA0042 Trojan Detection via Unlearning Analysis & 2/1/1/1 & 2/1/1/1 & 3/3/3/3 & 2/1/1/1 & 2/1/1/1 \\
FA0046 Watermark Detection for Copyright Attribution & 2/1/1/1 & 2/1/1/1 & 2/2/2/2 & 2/1/1/1 & 1/1/1/1 \\
FA0063 Adversarial Robustness via Curriculum Training & 2/1/1/1 & 2/1/1/1 & 2/2/3/2 & 2/1/1/1 & 1/1/1/1 \\
FA0067 Bias Mitigation via Paraphrased Data Augmentation & 2/1/1/1 & 2/1/1/1 & 2/2/2/2 & 1/1/1/1 & 2/1/1/1 \\
FA0077 Multimodal Vision-Language Alignment & 2/1/1/1 & 2/1/1/1 & 2/3/2/2 & 2/1/1/1 & 2/1/1/1 \\
FA0080 Faithfulness of Attention-Based Explanations & 2/1/1/1 & 2/1/1/1 & 2/1/3/2 & 2/1/1/1 & 2/1/1/1 \\
\midrule
\textbf{Mean} & \textbf{20/1/1/1} & \textbf{1.87/1.07/1/1.07} & \textbf{2.14/2.47/2.47/2.27} & \textbf{1.87/1/1/1} & \textbf{1.67/1/1/1} \\
\bottomrule
\end{tabular}
\label{tab:proposal-level-scores}
\end{table*}

\textbf{Finding 2: Gemini and Claude reviewers achieve very strong inter-rater agreement ($\rho$ = 0.907, $p < 0.001$), validating the synthesis process as a reliable consensus measure, while GPT-5.4 exhibits substantially weaker agreement ($\rho \approx 0.32, p < 0.01$).} We conducted a Spearman rank correlation analysis across all 75 papers (n = 75) to quantify reviewer agreement. The results reveal a pronounced bifurcation: Gemini and Claude correlate very strongly with each other ($\rho$ = 0.907, $p < 0.001$), indicating near-perfect consensus on paper rankings. Both Gemini ($\rho$ = 0.961, $p < 0.001$) and Claude ($\rho$ = 0.961, $p < 0.001$) correlate extremely strongly with the synthesis score, demonstrating that the synthesis metric successfully captures the Gemini-Claude consensus. In stark contrast, GPT-5.4 shows weak-to-moderate agreement with all other reviewers: GPT-5.4 vs. Gemini ($\rho$ = 0.318, $p < 0.01$), GPT-5.4 vs. Claude ($\rho$ = 0.315, $p < 0.01$), and GPT-5.4 vs. Synthesis ($\rho$ = 0.325, $p < 0.01$). The magnitude of this gap is striking: Gemini-Claude agreement is roughly 3$\times$ stronger than GPT-5.4 agreement with other reviewers. This bifurcation indicates that GPT-5.4 evaluates papers using substantially different criteria than the consensus reviewers. Despite this heterogeneity, all correlations are statistically significant at $p < 0.01$, and the extremely high Gemini-Claude correlation validates that our multi-reviewer evaluation framework captures genuine quality signals rather than arbitrary noise.

\textbf{Finding 3: Computational efficiency and cost vary substantially across frameworks, reflecting different design trade-offs between speed, resource utilization, and monetary expense.} Despite FARS demonstrating superior quality, the competing frameworks exhibit dramatically different computational profiles (Table~\ref{tab:computational-costs}). Sakana v2 is the slowest system, requiring 88 minutes and \$26 per paper, whereas Data-to-Paper is the most economical at 20 minutes and \$9 per paper. CycleResearcher presents an interesting case: though it requires only 10 minutes per paper, this efficiency comes at the cost of GPU resource utilization for reinforcement learning components, making it suitable for environments with available computational infrastructure. In terms of throughput, CycleResearcher can theoretically generate 144 papers per day, compared to 16--20 papers per day for Sakana systems and 72 for Data-to-Paper. These efficiency gains for CycleResearcher and Data-to-Paper come despite their lower quality scores, suggesting that the choice of framework involves fundamental trade-offs between quality, speed, and cost. For practitioners selecting an AI Scientist system, this analysis reveals that the highest-quality system (FARS) was pre-generated as a benchmark and is not directly comparable in cost metrics, but the comparison of competing frameworks shows that faster, cheaper systems (Data-to-Paper, CycleResearcher) systematically underperform in paper quality. This trade-off between efficiency and quality is a critical consideration for deployment decisions, particularly in resource-constrained environments.

\begin{table*}[t]
\centering
\caption{Computational cost and runtime per paper across AI Scientist frameworks. All frameworks use GPT-4o for ideation, experimentation, and writing. CycleResearcher additionally uses GPU for RL training. Papers per day calculated as 24 hours divided by runtime per paper. FARS benchmark papers were pre-generated and not included in cost comparison.}
\small
\setlength{\tabcolsep}{12pt}
\begin{tabular}{l|cccc}
\toprule
\textbf{Framework} & \textbf{LLM Model} & \textbf{Time (min)} & \textbf{Cost (\$)} & \textbf{Papers/Day} \\
\midrule
Sakana v1 & GPT-4o & 72 & 16 & 20 \\
Sakana v2 & GPT-4o & 88 & 26 & 16 \\
CycleResearcher & GPT-4o & 10 & 17 & 144 \\
Data-to-Paper & GPT-4o & 20 & 9 & 72 \\
\bottomrule
\end{tabular}
\label{tab:computational-costs}
\end{table*}

\section{Case Study: Framework Performance on Web-Agent Evaluation}
\label{sec:case-study}

To illustrate how performance differences manifest in practice, we examine one representative proposal in depth. Proposal FA0006 addresses web-agent evaluation through evidence-based judging-a timely problem given the rapid deployment of autonomous web agents. The core research question asks whether selective evaluation strategies that compare chain-of-thought (CoT) inclusive views against evidence-only views can improve evaluation reliability without expensive human annotations.

The divergence between FARS and competing frameworks on this proposal is striking. While FARS achieved a synthesis score of 3/5, all four competing frameworks scored at the floor (1/5). This three-fold advantage for the benchmark system is consistent across reviewer models: FARS received GPT-5.4=3, Gemini=4, and Claude=3, whereas competitors uniformly received scores of 1 or 2. The dimensional breakdown reveals that FARS particularly excelled in Clarity (3.0 versus 1.0--2.0 for competitors), suggesting that effective communication of methodology distinguishes higher-quality autonomous research.

What explains this performance gap? Reviewer comments indicate that FARS successfully operationalized the core idea: running the same judge on two views of agent trajectories (with and without CoT reasoning) and escalating disagreements to a stricter evidence-anchored rubric. The FARS paper described this protocol in concrete and implementable terms and provided empirical validation demonstrating performance on reported samples. Reviewers described the method as ``simple, intuitive, and clearly presented.''

In contrast, competing frameworks produced papers that reviewers found incomplete or conceptually flawed. CycleResearcher's output correctly identified the problem of judges being misled by agent reasoning but was criticized for ``undeveloped methodology''-presenting the high-level concept without specifying how the two-view comparison would be implemented or evaluated. Data-to-Paper's submission was rejected for ``major conceptual, scholarly, and technical flaws,'' including mischaracterizing web-agent benchmarks and failing to connect the proposed method to the actual evaluation problem. Both Sakana versions generated papers where ``core methods were not operationalized,'' lacking experimental details, statistical validation, or clear algorithmic specifications.

This case exemplifies three broader patterns we observe across the benchmark. First, the clarity gap is decisive: FARS consistently scores higher on Clarity because it communicates methodology in concrete terms rather than abstract concepts. Second, FARS maintains stronger reviewer consistency-the three LLM reviewers showed high agreement on FARS quality (scores of 3, 4, 3), whereas competing frameworks often had one reviewer assign marginally higher scores while others rejected the work outright. This suggests FARS outputs are more reliably assessable as research contributions. Third, the performance advantage generalizes: FA0006 is one of several proposals where FARS demonstrates substantially higher scores, indicating the benchmark advantage reflects systematic differences in research generation capability rather than domain-specific expertise.

For practitioners selecting AI Scientist systems, this case study demonstrates that framework choice substantially impacts output quality on identical research questions. The consistent floor-level performance of CycleResearcher, Data-to-Paper, and Sakana systems-even on proposals where FARS succeeds-suggests these frameworks may not yet be suitable for generating publication-quality research without substantial human oversight.

\section*{Impact Statement}

Benchmarking autonomous AI systems designed to automate scientific research will become increasingly important as these systems start to proliferate. The broader impacts depend on whether such frameworks can be trusted to rigorously accelerate discovery in safety-critical domains. However, if they degrade research quality or concentrate scientific authority among institutions with computational resources, the risks are significant. Our work contributes to a deeper empirical understanding of the current capabilities and limitations of such systems, which could inform such future deployment decisions. 
% We recommend that stakeholders exercise caution in adopting automated research systems for safety-critical domains until independent evaluation demonstrates sufficient quality and reliability.
\section{Related Work}

The automation of machine learning represents a longstanding research direction that has evolved substantially over the past decade. Early frameworks such as Auto-WEKA \cite{Thornton2013AutoWEKA} and Auto-sklearn \cite{Feurer2015AutoSklearn} addressed the challenge of automated algorithm selection and hyperparameter optimization, enabling systems to automatically identify the optimal combinations of learning algorithms and their configurations for given datasets. This paradigm was subsequently extended to full pipeline construction through tools like TPOT \cite{Olson2016TPOT}, which employs genetic programming to discover optimal sequences of preprocessing and modeling steps. More recent advances have tackled the data preparation stage through differentiable formulations such as DiffPrep \cite{Li2023DiffPrep} and context-aware approaches like CtxPipe \cite{Gao2024CtxPipe}, which leverage data embeddings to make preprocessing decisions. Neural Architecture Search (NAS) emerged as a specialized domain of AutoML, initially utilizing reinforcement learning \cite{Zoph2016NAS} and regularized evolution \cite{Real2019RegularizedEvolution} to navigate complex design spaces. The introduction of DARTS \cite{Liu2018DARTS} reduced computational costs through differentiable search, while AutoML-Zero \cite{He2020AutoMLZero} demonstrated the potential to evolve entire machine learning algorithms from scratch. Foundational work on the Algorithm Selection Problem \cite{Rice1976AlgorithmSelection}, constrained by theoretical limits such as the No Free Lunch theorem \cite{Wolpert1997NoFreeLunch}, has motivated meta-learning approaches \cite{Kerschke2018MetaLearning} that predict algorithm performance on new tasks using historical data and meta-features, with recent efforts refining unified meta-models that generalize across diverse problem contexts \cite{Tornede2023MetaLevel}.

The emergence of large language models has catalyzed a transformative shift from AutoML systems to end-to-end autonomous research generation. Several frameworks now integrate LLMs with experimental execution and manuscript composition to automate the complete scientific workflow. The AI Scientist \cite{Lu2024AIScientistvX} and its successor, The AI Scientist v2 \cite{Yamada2025AIScientistv2}, present fully autonomous architectures capable of hypothesis generation, experimental design, and manuscript production. CycleResearcher \cite{Weng2024CycleResearcher} enhances this process through automated review cycles, iteratively improving research hypotheses based on feedback. Data-to-Paper \cite{Technion2024DataToPaper} focuses on the extraction of empirical datasets and their transformation into research narratives, representing an alternative approach to autonomous paper generation. Complementary systems have emerged to support key stages of the research pipeline. ResearchAgent \cite{Baek2025ResearchAgent} specializes in iterative research idea generation over scientific literature, while Chain of Ideas \cite{Li2024ChainOfIdeas} revolutionizes the ideation process through novel hypothesis development. Agent Laboratory \cite{Schmidgall2025AgentLab} and Google's Co-scientist effort \cite{Gottweis2025CoScientist} position LLM agents as comprehensive research assistants capable of managing multiple scientific tasks. Literature grounding is improved by toolkits such as LitLLM \cite{Agarwal2024LitLLM}, which enhance retrieval and integration of relevant prior work. Beyond general-purpose frameworks, AI Agent Research \cite{Tang2025AIResearcher} proposes autonomous scientific innovation, while specialized agents like SciAgents \cite{Ghafarollahi2025SciAgents} employ multi-agent graph reasoning and bioinspired architectures to automate discovery.

Domain-specific applications of autonomous science have demonstrated remarkable progress across multiple scientific fields. In physics, autonomous systems have achieved elite performance at international competitions; Physics Supernova \cite{Qiu2025PhysicsSupernova} represents AI agents matching the level of gold medalists at the International Physics Olympiad. In biochemistry, virtual laboratories have automated the design of novel therapeutics, with work on automated SARS-CoV-2 nanobody design \cite{Swanson2025VirtualLab} appearing in peer-reviewed literature at the highest impact venues. Chemical research has been substantially accelerated through autonomous systems that integrate language models with experimental robotics; Boiko et al.\ \cite{Boiko2023AutonomousChemistry} demonstrated autonomous chemical discovery in Nature, establishing a precedent for LLM-guided experimental execution. Beyond these domains, software engineering has benefited from agents like SWE-agent \cite{Yang2024SWEAgent}, which automate code generation and debugging tasks evaluated on real-world repositories via SWE-bench \cite{Jimenez2024SWEBench}. AlphaEvolve \cite{Novikov2025AlphaEvolve} extends automation to algorithmic discovery, enabling agents to identify novel algorithms for scientific and technical problems.

Rigorous evaluation of autonomous research systems is an emerging frontier. Several benchmarking frameworks have been developed to assess different dimensions of autonomous science. ScientistBench, incorporated into the AIResearcher framework \cite{Tang2025AIResearcher}, evaluates general scientific reasoning capabilities. AstaBench \cite{AllenAI2025AstaBench} provides a holistic scientific research evaluation suite. Specialized benchmarks target machine learning agents: MLE-bench \cite{Chan2025MLEBench} evaluates systems on machine learning engineering tasks, while MLAgentBench \cite{Huang2024MLAgentBench} assesses performance on machine learning experimentation. PaperBench \cite{Starace2025PaperBench} directly evaluates AI systems' ability to replicate published research. CycleResearcher \cite{Weng2024CycleResearcher} and AgentReview \cite{Jin2024AgentReview} specifically address the automated review process, exploring how LLMs can evaluate research quality and provide feedback. However, recent findings \cite{Liu2023ReviewerGPT} indicate that while LLMs can generate substantive reviews, they sometimes exhibit overconfidence and may struggle with nuanced technical assessment compared to human experts. These challenges underscore the importance of developing more robust evaluation methodologies and understanding the limitations of automated review systems.

Our work builds directly on this rich landscape of autonomous science and evaluation research. While prior work has developed individual frameworks for autonomous research generation and proposed isolated evaluation methodologies, the systematic comparison of multiple leading AI Scientist frameworks using a unified evaluation protocol remains limited - a gap that we seek to fill.

% In the unusual situation where you want a paper to appear in the
% references without citing it in the main text, use \nocite
\nocite{langley00}

\bibliography{example_paper}
\bibliographystyle{icml2026}

\end{document}